\pdfoutput=1

\documentclass[11pt]{article}

\usepackage[final]{acl}

\usepackage{times}
\usepackage{latexsym}
\usepackage{booktabs}

\usepackage[most]{tcolorbox}

\usepackage[T1]{fontenc}

\usepackage[utf8]{inputenc}

\usepackage{microtype}

\usepackage{inconsolata}

\usepackage{graphicx}

\title{DRISHTIKON: A Multimodal Multilingual Benchmark for Testing Language Models' Understanding on Indian Culture}

\author{
  \textbf{Arijit Maji\textsuperscript{1}}, 
  \textbf{Raghvendra Kumar\textsuperscript{1}}, 
  \textbf{Akash Ghosh\textsuperscript{1}}, 
  \textbf{Anushka\textsuperscript{2}}, 
  \textbf{Nemil Shah\textsuperscript{3}}, \\
  \textbf{Abhilekh Borah\textsuperscript{4}}, 
  \textbf{Vanshika Shah\textsuperscript{5}}, 
  \textbf{Nishant Mishra\textsuperscript{1}}, 
  \textbf{Sriparna Saha\textsuperscript{1}} \\
  \\
  \textsuperscript{1} Indian Institute of Technology Patna, India \\
  \textsuperscript{2} Banasthali Vidyapeeth University, Rajasthan, India \\
  \textsuperscript{3} Pandit Deendayal Energy University, India \\
  \textsuperscript{4} Manipal University Jaipur, India \\
  \textsuperscript{5} Dwarkadas J. Sanghvi College of Engineering, India \\
  \\
  \small{\textbf{For inquiries:}} \\
  \small{\{\texttt{arijit\_2311ai25, raghvendra\_2221cs27, akash\_2321cs19, nishant\_2312res420, sriparna}\}@iitp.ac.in} \\
  \small{\{\texttt{guptaanushka024, nemilshah212005, abhilekhkey, vanss2808}\}@gmail.com}
}

\begin{document}
\maketitle
\begin{abstract}
We introduce DRISHTIKON, a first-of-its-kind multimodal and multilingual benchmark centered exclusively on Indian culture, designed to evaluate the cultural understanding of generative AI systems. Unlike existing benchmarks with a generic or global scope, DRISHTIKON offers deep, fine-grained coverage across India’s diverse regions, spanning 15 languages, covering all states and union territories, and incorporating over 64,000 aligned text-image pairs. The dataset captures rich cultural themes including festivals, attire, cuisines, art forms, and historical heritage amongst many more. We evaluate a wide range of vision-language models (VLMs), including open-source small and large models, proprietary systems, reasoning-specialized VLMs, and Indic-focused models, across zero-shot and chain-of-thought settings. Our results expose key limitations in current models’ ability to reason over culturally grounded, multimodal inputs, particularly for low-resource languages and less-documented traditions. DRISHTIKON fills a vital gap in inclusive AI research, offering a robust testbed to advance culturally aware, multimodally competent language technologies.The Dataset and inferencing codes are publicly available at \footnote{{\url{https://tinyurl.com/DrishtikonDataset}}}.
\end{abstract}


\section{Introduction}

Language Models (LMs) and their multimodal successors have revolutionized natural language processing, excelling at tasks like generation, retrieval, translation, summarization and reasoning \cite{10.5555/3495724.3495883,devlin-etal-2019-bert,ghosh2024healthalignsumm,ghosh2024medsumm,ghosh2024sights,kumar2023diving,verma-etal-2023-large,kumar-etal-2025-cosmmic}. With the advent of Large Multimodal Models (LMMs) \cite{ghosh2024exploring}, Indic-centric Language Models (ILMs) \cite{jain2020indic,kumar2024indicbart}, and parameter-efficient Small Language Models (SLMs),
AI systems are now increasingly deployed across global and multilingual domains \cite{10.5555/3600270.3602281,ghosh2025multilingual}. 

However, despite their linguistic fluency, these models remain largely blind to the socio-cultural richness that defines real-world communication \cite{maji-etal-2025-sanskriti}. Especially in culturally rich regions like India, with its diverse traditions, languages, rituals, clothing, festivals, and food, models often struggle, either misinterpreting, oversimplifying, or overlooking the context needed for culturally aware reasoning \cite{blodgett-etal-2020-language}. This presents serious concerns for AI systems deployed in education, governance, healthcare, heritage documentation, and creative industries, where cultural misalignment can lead to misinformation, bias amplification, and exclusion \cite{liang2022holistic}.

\textbf{Research Gap:} Existing benchmarks mainly target linguistic generalization (e.g., TyDi QA \cite{clark-etal-2020-tydi}, XQUAD \cite{artetxe-etal-2020-cross}) or basic image-text alignment, but lack the cultural specificity and multimodal grounding needed for culturally competent AI. They often miss region-specific symbolism such as the spiritual role of Baul music in Bengal, Warli iconography in Maharashtra, or Manipuri dance attire. Furthermore, efforts like CVQA \cite{romero2025cvqa}, World Value Survey-based benchmarks \cite{yadav2025beyond,durmus2023towards}, ALM \cite{vayani2024all}, and CulturalBench \cite{chiu2024culturalbench} fall short of offering a holistic framework for Indian cultural diversity. They do not jointly cover multiple Indian languages, rich visual modalities, and nuanced cultural understanding. Crucially, none span all states and union territories, limiting their value for culturally grounded AI evaluation in India.

\textbf{Research Motivation:} To address this gap, we introduce \textbf{DRISHTIKON}, a \textbf{multimodal, multilingual benchmark} dedicated to Indian culture. It evaluates vision-language models’ ability to reason over culturally grounded content by aligning text and visuals. DRISHTIKON spans 15 Indian languages (including English), covers all 28 states and 8 union territories, and comprises 64,288 carefully curated instances reflecting traditional arts, festivals, attire, architecture, cuisine, and regional practices. We further benchmark several state-of-the-art vision-language models (VLMs), including open-source small models (\textit{e.g., SmolVLM-256M-Instruct \cite{marafioti2025smolvlm}, InternVL3-1B \cite{iv1,iv2,iv3,iv4}}), large VLMs (\textit{e.g., Janus-Pro-7B \cite{janus}, Qwen2-VL-7B-Instruct \cite{qw1,qw2}, Llama-4-Scout-17B-16E-Instruct \cite{meta2025llama4scout}, LLaVA-1.6-Mistral-7B \cite{liu2023improved}, InternVL3-14B \cite{iv1,iv2,iv3,iv4}, Gemma-3-27B-IT \cite{team2025gemma}, Qwen2.5-Omni-7B \cite{xu2025qwen2}}), proprietary systems (\textit{e.g., GPT-4o-mini \cite{achiam2023gpt}
}), reasoning-specialized VLMs (\textit{e.g., Kimi-VL-A3B-Thinking \cite{kimi}}), and Indic-aligned models (\textit{e.g., Chitrarth \cite{khan2024chitrarth}, Maya \cite{maya}}), under zero-shot, and chain-of-thought (CoT) prompting paradigms \cite{sahoo2024systematic}. To encapsulate our work, we highlight the following key contributions:

\emph{
\begin{itemize}
    \item We introduce \textbf{DRISHTIKON}, the first \textit{multimodal, multilingual} benchmark tailored to evaluate cultural reasoning in India, spanning 15 languages and capturing diverse visual-textual expressions from all 28 states and 8 union territories.
    \item The dataset comprises 64,288 carefully curated instances enriched with fine-grained annotations and multiple-choice questions, enabling systematic assessment of perception, inference, and cultural alignment in VLMs.
    \item We conduct a large-scale evaluation of state-of-the-art VLMs, including compact, large-scale, proprietary, reasoning-specialized, and Indic-aligned models, under zero-shot and chain-of-thought prompting settings.
    \item DRISHTIKON unveils critical gaps in VLM performance on culturally grounded tasks, especially in low-resource languages and region-specific contexts, underscoring the need for culturally inclusive AI development.
\end{itemize}
}

\section{Related Works}

To contextualize our contribution, we structure the related work into two key areas: (i) multimodal and multilingual cultural benchmarks, and (ii) datasets centered on culturally grounded Indian languages and regional diversity.

\subsection{Multimodal and Multilingual Cultural Benchmarks}

\cite{schneider2024m5} proposed M5, a benchmark for vision-language tasks across 41 languages, showing that larger models often underperform smaller ones in low-resource settings. \cite{romero2025cvqa} introduced CVQA, a culturally diverse VQA benchmark with 9k questions in 26 languages from 28 countries, exposing model limitations in cultural and linguistic diversity. \cite{schneider2025gimmick} presented GIMMICK, spanning 728 cultural facets across 144 countries, where evaluations of LVLMs and LLMs revealed strong Western bias and weak cultural understanding. Similarly, \cite{nayak-etal-2024-benchmarking} introduced CulturalVQA to evaluate geo-diverse reasoning across 11 countries, revealing that GPT-4V and Gemini performed better on North American contexts while struggling with African cultural content.

Other relevant efforts include ALM \citep{vayani2024all}, Blend \citep{myung2024blend}, GlobalBench \citep{singh2024global}, SEA-Eval \citep{wang2024seaeval}, CUBE \citep{senthilkumar2024beyond}, World Wide Recipe \citep{magomere2025worldwiderecipecommunitycentred}, IndoCulture \citep{koto2024indoculture} and MultiLoKo \citep{hupkes2025multiloko}, which address linguistic or regional diversity. Region-specific studies, such as JMMMU \citep{onohara2024jmmmu}, focus on Japanese multimodal understanding. However, \emph{none of these benchmarks offer the fine-grained, culturally rich, and linguistically broad coverage of India that our work uniquely provides}.

\begin{figure*}[htbp]
    \centering
    \includegraphics[width=0.95\textwidth]{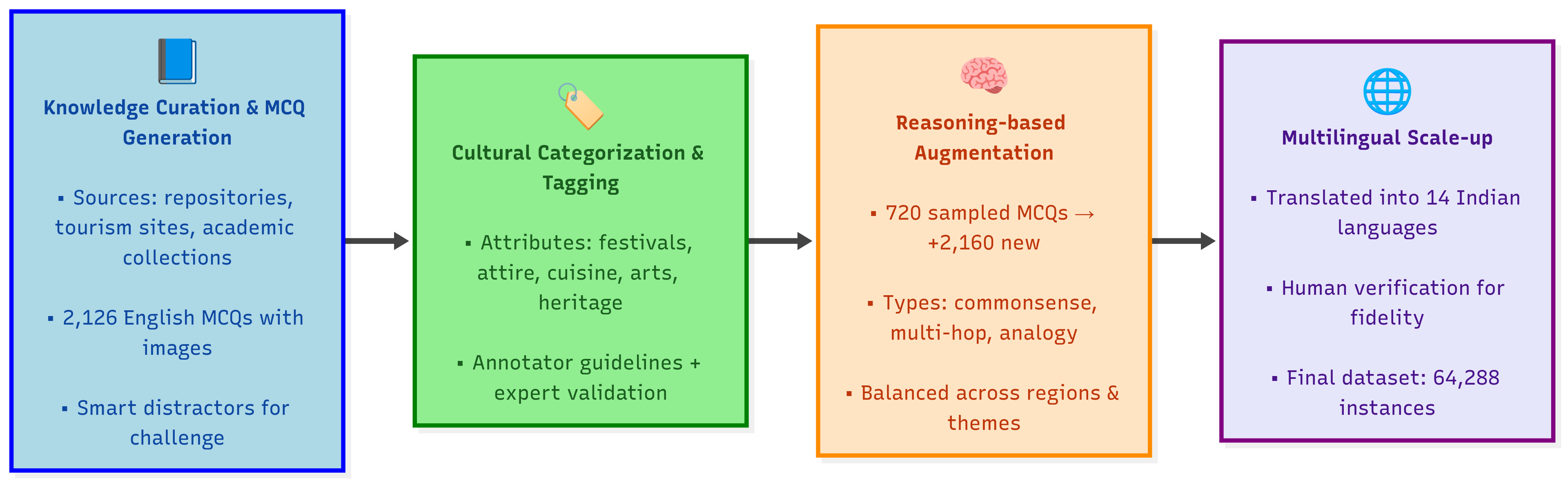}
    \caption{DRISHTIKON dataset creation pipeline showing knowledge curation, MCQ generation, cultural categorization, reasoning-based augmentation, multilingual translation, and final dataset assembly.}
    \label{fig:dataset_pipeline}
\end{figure*}

\subsection{Regionally and Culturally Rich Indian Corpora}
Prior efforts have developed diverse Indian corpora addressing language, culture, and social biases. \cite{seth2024dosa} introduced DOSA, a community-driven dataset of 615 artifacts from 19 subcultures, revealing LLM performance disparities. \cite{sahoo2024indibias} presented IndiBias, a bilingual dataset highlighting caste, religion, and gender biases. \cite{kakwani2020indicnlpsuite} and \cite{doddapaneni-etal-2023-towards} released large-scale Indic corpora (8.8B and 20.9B tokens) with resources like IndicGLUE and IndicXTREME to advance Indic NLP. \cite{bhatt2022cultural} proposed frameworks for NLP fairness, while \cite{hasan2024nativqa} developed NativQA and MultiNativQA to fine-tune models for low-resource, dialect-rich languages.

For factual and culturally grounded QA, \cite{rohera2024l3cube} created L3Cube-IndicQuest with 200 QA pairs across English and 19 Indic languages in five domains. \cite{10.1145/3677525.3678666} introduced Indian-BhED, exposing LLMs’ stereotypical outputs and underscoring the need for culturally diverse fairness evaluations. For broader context, see surveys and studies such as \citep{maji-etal-2025-sanskriti,pawar2024surveyculturalawarenesslanguage,adilazuarda-etal-2024-towards,kharchenko2024well,karinshak2024llm,alkhamissi2024investigating,rystrom2025multilingual,shen-etal-2024-understanding,winata2025worldcuisines}.

\emph{While prior work addressed sociolinguistic bias, dialects, or factual QA in India, our study uniquely integrates multilingual, multimodal, and culturally grounded question answering, emphasizing visual reasoning across all 36 states and union territories. DRISHTIKON is the first large-scale benchmark to holistically evaluate cultural competence in generative models using both text and visuals.}

\section{DRISHTIKON Dataset Construction}
Figure~\ref{fig:dataset_pipeline} illustrates the complete DRISHTIKON dataset creation pipeline, from knowledge curation and MCQ generation to reasoning-based augmentation and multilingual scaling.

\subsection{Knowledge Curation and MCQ Generation}
We curated a rich knowledge base capturing India’s diverse socio-cultural landscape using authoritative sources such as national repositories, state tourism portals, academic collections, and curated crowd-sourced platforms. Content spans festivals, attire, cuisines, folk traditions, monuments, personalities, and more (full details in Appendix~\ref{app-1}).

Inspired by vision-language QA datasets (e.g., CVQA~\cite{romero2025cvqa}) and cultural evaluations like DOSA~\cite{seth2024dosa}, we framed multiple-choice questions (MCQs) with one correct answer and three distractors. The 4-option format balances cultural granularity, annotator effort, and model load, while aligning with prior benchmarks (CVQA~\cite{romero2025cvqa}, CulturalVQA~\cite{nayak-etal-2024-benchmarking}). Though more distractors were possible, they risked diluting focus and raising annotation costs, especially across 64k+ instances. The 4-choice setup also lowers chance-level guessing (25\%) and enables consistent evaluation. 

\begin{figure*}[htbp]
    \centering
    \includegraphics[width=0.90\textwidth]{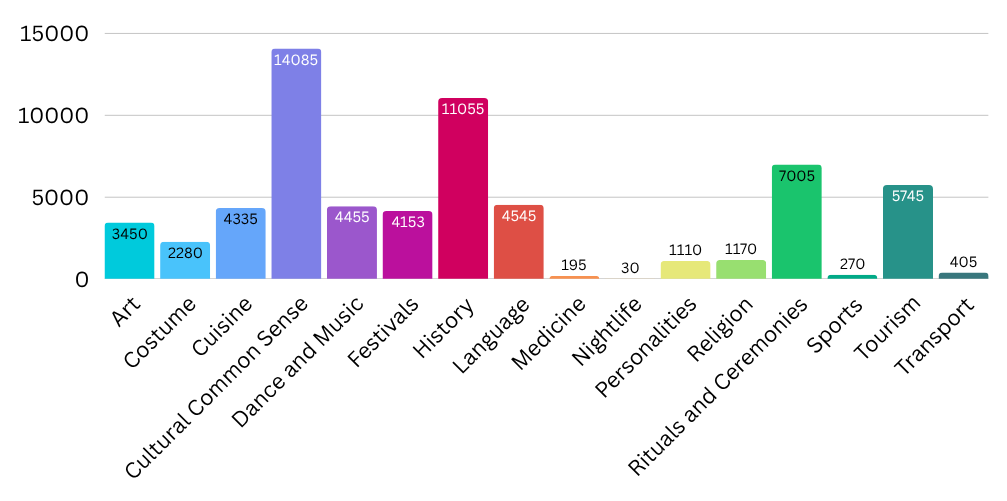} 
    \caption{Question distribution across cultural aspects.}
    \label{fig:cultural_aspects_distribution}
\end{figure*}

Distractors were generated through a semi-automated process followed by human curation, ensuring diversity in semantic proximity. Some distractors were intentionally close to the correct answer (e.g., from the same state or cultural category) to test fine-grained knowledge and distractor resistance. Others were thematically plausible but incorrect (e.g., attire from a neighboring region). To avoid uniform similarity, each MCQ typically contained a controlled mix, one semantically close distractor, one reflecting a popular misbelief, and one unrelated but superficially similar option. 

We authored 2,126 English MCQs, proportionally covering all 28 states and 8 union territories, emphasizing cultural significance while avoiding stereotypes or trivia.

Each MCQ underwent two-pass validation for factual accuracy, clarity, and cultural sensitivity. To support multimodal comprehension, every question was paired with a culturally relevant image, selected via controlled Google Image Search for clarity, contextual fit, and inclusivity. While open-ended formats can test deeper reasoning, we opted for MCQs to ensure comparability and reproducibility: they provide a consistent evaluation signal and allow robust accuracy-based scoring across 15 languages and multiple VLMs, including those with constrained generation capabilities. We further mitigated potential ``test-taking strategies” by designing semantically rich and reasoning-augmented MCQs, such as multi-hop, analogy-based, and common-sense cultural formats, that resist superficial pattern matching. Nonetheless, we acknowledge the merit of open-ended formats and plan to incorporate them in future expansions of DRISHTIKON for joint evaluation across free-text generation and MCQ reasoning paradigms. Annotation methodology, agreement metrics, and adjudication procedures are outlined in Appendix~\ref{app-2}.

\subsection{Cultural Categorization and Attribute Tagging}
For structured cultural benchmarking, each question-image pair was annotated with one or more high-level cultural attributes. These attributes emerged from a dynamic taxonomy designed to reflect India’s cultural diversity. While the taxonomy is still evolving, initial categories include aspects such as \textit{attire}, \textit{festivals}, \textit{cuisine}, \textit{rituals}, \textit{folk arts}, \textit{heritage sites}, and \textit{notable personalities}, among others. Definitions and taxonomy details are available in Appendix~\ref{app-4}.

Manual tagging was performed by trained annotators using standardized guidelines to maintain consistency. Ambiguities such as multi-attribute questions or overlapping cultural references were resolved through consensus meetings and expert adjudication. Detailed attribute statistics, overlap patterns, and edge case examples are presented in Appendix~\ref{app-4}. This structured labelling enabled targeted slicing of the dataset and supported fine-grained evaluation of model performance across cultural modalities.

\begin{figure}[htbp]
    \centering
    \includegraphics[width=0.5\textwidth]{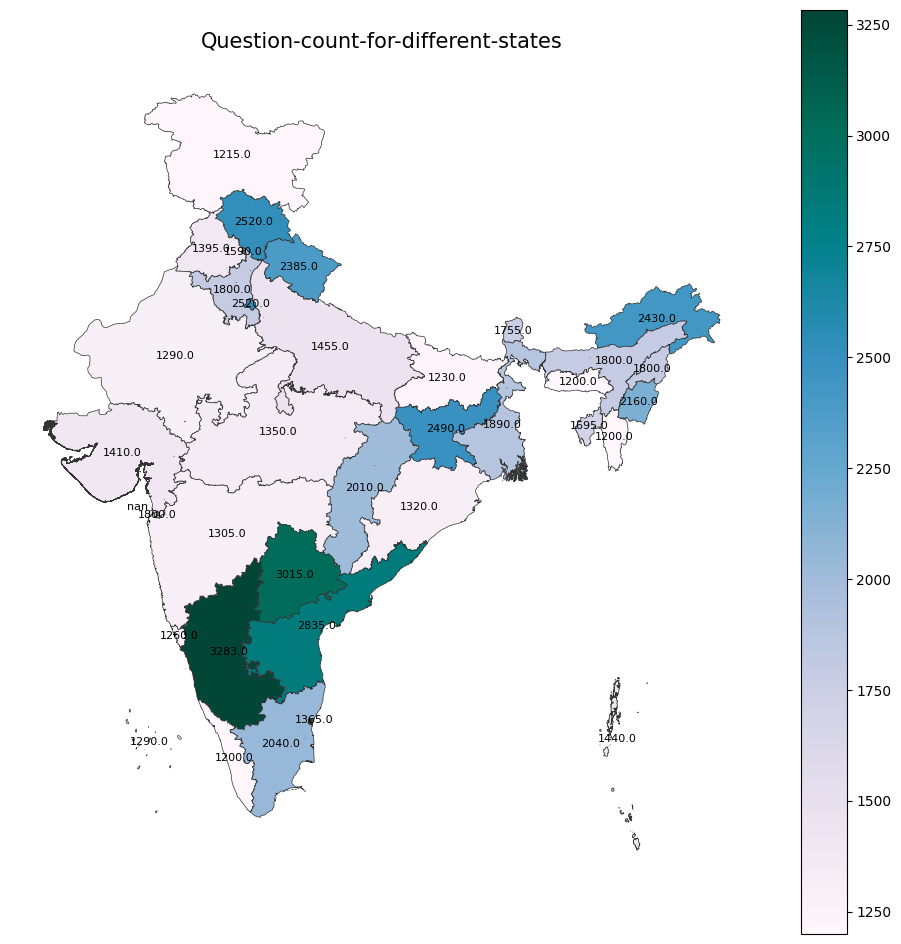} 
    \caption{State-wise and Union Territory-wise question distribution.}
    \label{fig:statewise_distribution}
\end{figure}

Figure~\ref{fig:cultural_aspects_distribution} shows the thematic distribution of questions across cultural attributes, while Figure~\ref{fig:statewise_distribution} visualizes their geographic spread across India.  \textbf{Further Dataset Statistics:} Due to space constraints, we defer comprehensive statistical breakdowns, including question-type frequencies (factual, reasoning, analogy), and detailed state/UT questions coverage to the appendix. Appendix~\ref{app-8} presents visualizations and tables to support deeper analysis.

\subsection{Reasoning-based Question Augmentation}
To move beyond surface recognition and test deeper inferencing, we introduced reasoning-based question augmentation. From the original 2,126 MCQs, a balanced subset of 720 ($\approx$ 20 per region) was selected to ensure equitable regional representation. These were augmented into three reasoning categories: \textbf{Common Sense Cultural}, requiring everyday cultural inference (e.g., attire or food pairing); \textbf{Multi-hop Reasoning}, combining multiple cultural aspects (e.g., linking a dance form to its festival and state); and \textbf{Analogy}, framing cultural pattern-matching (e.g., relating dishes or art forms across states).

The 20-question cap per region was based on the lowest available count, ensuring uniform augmentation and balanced evaluation. For regions with more data, stratified sampling captured diverse cultural themes (attire, cuisine, festivals, heritage), mitigating bias. All augmented questions were manually reviewed for logic, relevance, and fluency, resulting in 2,160 additional MCQs with greater inferential depth while preserving regional and thematic balance. Detailed sampling methods, example transformations, and culturally grounded chain-of-thought prompts are provided in Appendices~\ref{app-5} and~\ref{Q-types}, with CoT details in Appendix~\ref{cot}.

\subsection{Multilingual Translation and Dataset Scale-up}

To reflect the linguistic diversity of India and promote inclusive model evaluation, we extend DRISHTIKON into a multilingual benchmark. All 2,126 base questions and 2,160 reasoning-augmented MCQs were translated into 14 Indian languages: Hindi, Bengali, Tamil, Telugu, Marathi, Kannada, Malayalam, Gujarati, Punjabi, Odia, Assamese, Urdu, Konkani, and Sindhi. This enables fine-grained analysis of language-specific generalization and cultural grounding.

We utilized the Gemini Pro \cite{deepmind2025geminipro} language model for translation, motivated by its demonstrated strengths in multilingual semantic fidelity and cultural contextualization, as evidenced by recent evaluations on FLORES-200 and XTREME-UP benchmarks~\cite{nllb2022,ref1,ref2}. In addition to its high translation quality, Gemini Pro offered practical scalability for processing a dataset of this magnitude. To mitigate risks of hallucination or mistranslation\cite{sahoo2024unveiling}, we adopted a two-stage human verification protocol on stratified samples, assessing translations for meaning preservation, fluency, and cultural relevance. For terms lacking direct equivalents in target languages, such as region-specific food items or artistic forms, transliteration or adaptive context-sensitive phrasing was applied. Language-specific challenges and resolutions are detailed in Appendix~\ref{app-6}.
The resulting dataset comprises 64,288 question-image-language triples spanning 36 regions, 16 cultural themes, and multiple question types. Each instance includes a culturally grounded question, four answer options with one correct label, the associated image URL (path once downloaded), and structured tags such as the question type, language, state/UT, and cultural attribute. The dataset is provided in tabular sheet (excel,csv) format for ease of use and analysis.

Together, these choices make DRISHTIKON the first large-scale, multilingual, multimodal benchmark explicitly designed to evaluate cultural competence and generalization in generative AI systems. Detailed information regarding the annotator distribution, qualifications, training, and compensation is provided in Appendix~\ref{annotators}.

\begin{figure*}[htbp]
\centering
\includegraphics[width=\textwidth]
{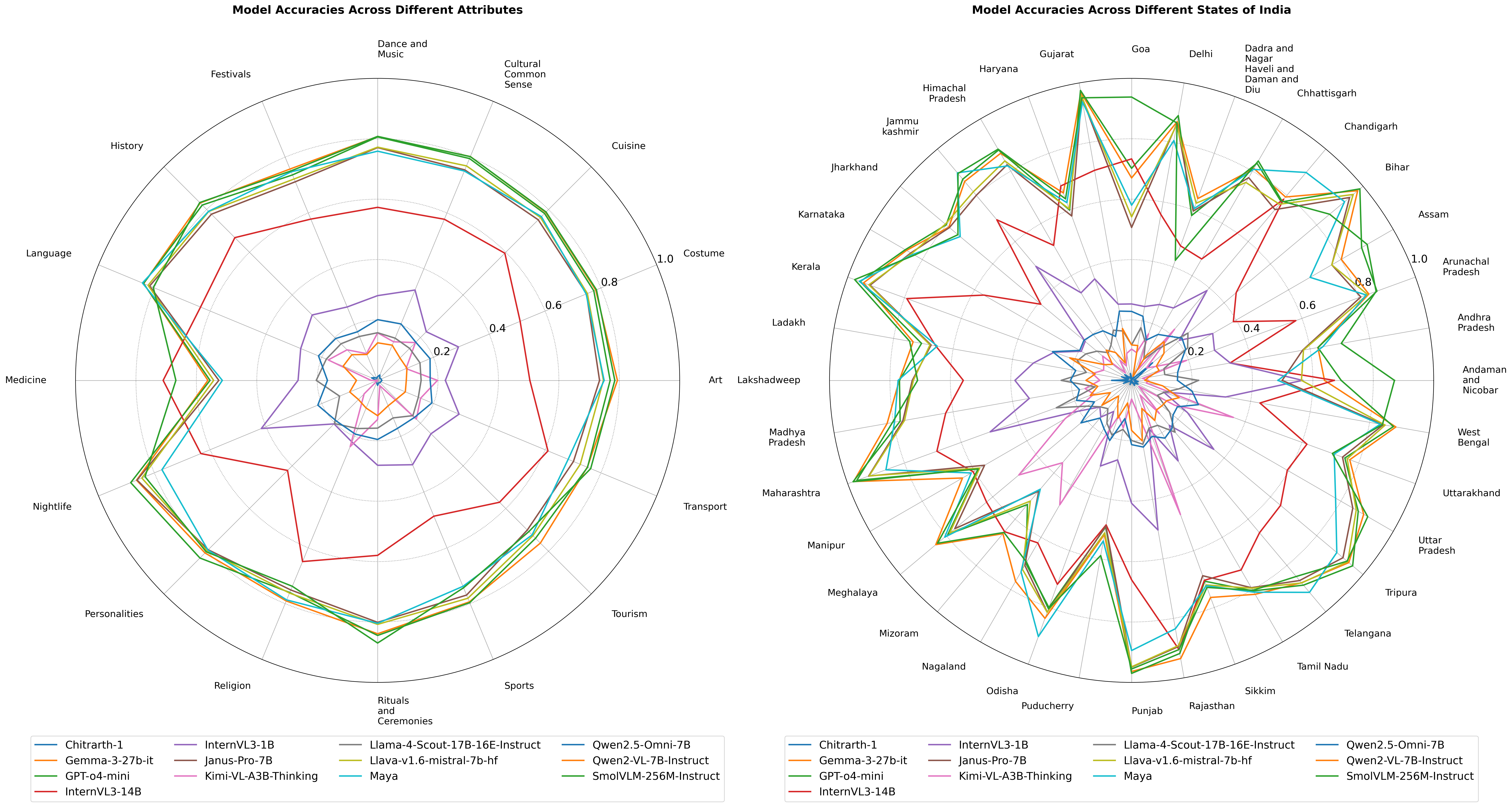}
\caption{Combined spider graph showing accuracy distribution across cultural attributes (left) and Indian states/UTs (right). This visualization highlights both thematic and regional performance variations across the evaluated vision-language models.}
\label{fig:combined_spider_graph}
\end{figure*}

\section{Experimental Setup}

To ensure a fair comparison across diverse vision-language models (VLMs), we adopt a unified evaluation protocol wherever possible. We standardize inputs with image resolutions of $224 \times 224$ or higher (depending on model capacity), and apply prompt templates consistent with each model's instruction tuning. The maximum token length is set based on the architecture-specific constraints, allowing multi-turn reasoning when supported. The hyperparameter settings for each model are detailed in Appendix~\ref{app:hyperparams}.

\textbf{Models:} We evaluate a wide range of vision-language models (VLMs), spanning multiple scales and capabilities. These include open-source small models such as SmolVLM-256M-Instruct and InternVL3-1B; large-scale models like Janus-Pro-7B, Qwen2-VL-7B-Instruct, Llama-4-Scout-17B-16E-Instruct, LLaVA-1.6-Mistral-7B, InternVL3-14B, Gemma-3-27B-IT, and Qwen2.5-Omni-7B; proprietary systems like GPT-4o-mini; reasoning-specialized models such as Kimi-VL-A3B-Thinking; and Indic-aligned models including Chitrarth and Maya. \textbf{Accuracy} is used as the primary evaluation metric, reflecting the proportion of correctly answered multiple-choice questions.

\section{Results}

\begin{figure*}[htbp]
\centering
\includegraphics[width=\textwidth, trim={0cm 1cm 0cm 0cm}, clip]{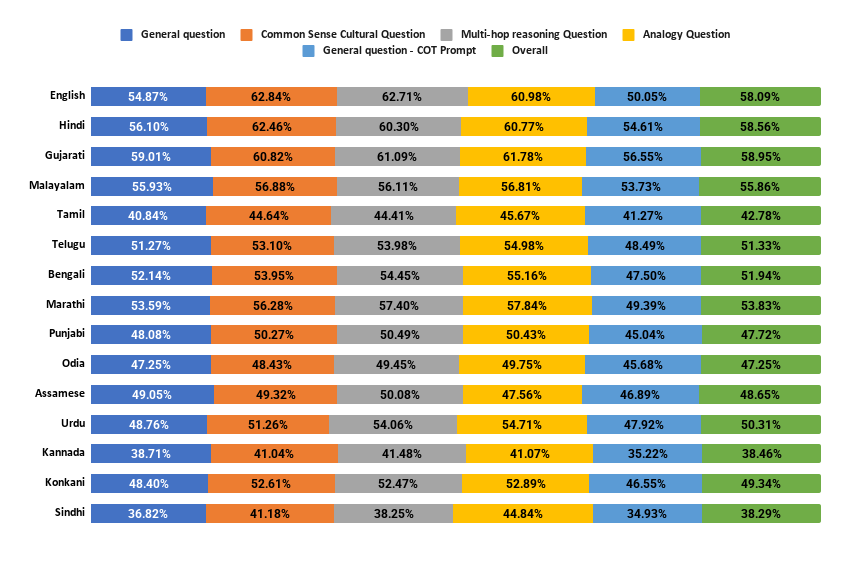}
\caption{Accuracy across languages under different question-type settings. Each percentage indicates the average accuracy (aggregated over all evaluated models) for a specific language–question type pair.}
\label{fig:accuracy_zero_shot_languages}
\end{figure*}

\begin{figure*}[htbp]
\centering
\includegraphics[width=\textwidth]{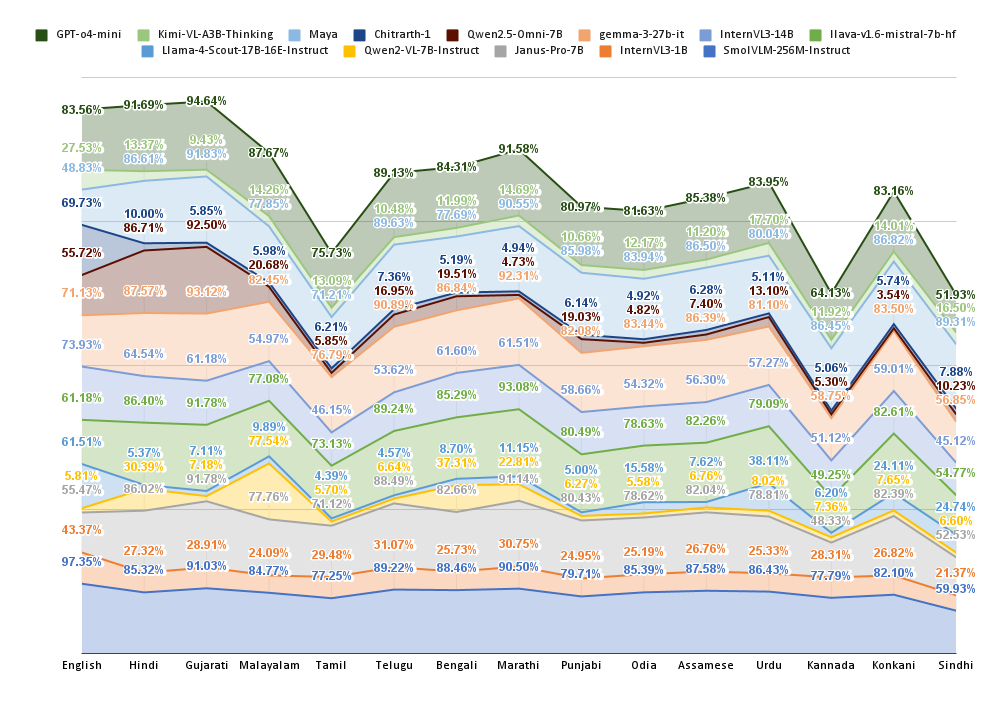}
\caption{Accuracy of each model across different languages, highlighting multilingual performance variations. Reported percentages represent the average accuracy for each language–model pair over the entire dataset.}
\label{fig:accuracy_zero_shot_models}
\end{figure*}

In this section, we present the evaluation results of multiple vision-language models (VLMs) on the DRISHTIKON dataset. We assess the models' performance across 15 Indian languages (English inclusive) and across various question types. The results are visualized through several illustrations (Figures. \ref{fig:combined_spider_graph}, \ref{fig:accuracy_zero_shot_languages} \& \ref{fig:accuracy_zero_shot_models}), offering insights into accuracy trends across cultural attributes, regional distributions, languages, and models.


\subsection{Analysis of Radar Graphs}

The radar plots in Figure~\ref{fig:combined_spider_graph} offer a comprehensive view of how vision-language models engage with culturally grounded attributes and geographically anchored knowledge. Models exhibiting broad and uniform radial coverage signal a robust alignment between visual and linguistic modalities, likely resulting from exposure to diverse, multimodal training data. Their smooth contours reflect an ability to generalize across both concrete cultural elements, such as attire, cuisine, and festivals, and more nuanced attributes like language, heritage, or environment. In contrast, models with jagged or constricted profiles reveal gaps in cultural grounding, particularly with abstract or context-dependent concepts like religion, nightlife, or medicine, which demand deeper socio-cultural and inferential reasoning.

Similarly, the radar plot of model accuracies across Indian states illustrates how well these models internalize region-specific cues. States with strong media presence or distinct cultural signatures, such as Kerala, Gujarat, and West Bengal, show higher and more consistent performance, hinting at the role of representation in pretraining corpora. Meanwhile, smaller or less represented regions like Lakshadweep, Mizoram, and Dadra and Nagar Haveli see lower accuracies, exposing geographic biases and uneven regional learning. 

Notably, even the best-performing models show fluctuations across states, underscoring persistent challenges in capturing India's cultural and linguistic diversity. Together, these radar charts reveal not just performance disparities but also hidden weaknesses, reinforcing the need for culturally inclusive, geographically balanced fine-tuning to ensure equitable and context-aware multimodal understanding.

\subsection{RQ1: To what extent does model scale correlate with performance in multilingual multimodal tasks?}

\textbf{Answering RQ1: Model-wise Performance Insights}
Among the evaluated models, proprietary large language models such as \texttt{GPT-4o mini} consistently deliver top-tier performance across all languages and question types, reflecting the advantage of extensive instruction tuning and large-scale vision-language alignment. Furthermore, \texttt{Maya}, despite being regionally focused and relatively lightweight, demonstrates competitive accuracy, challenging the assumption that scale alone drives multilingual multimodal performance. Following closely are SLMs such as \texttt{SmolVLM-256M-Instruct} and \texttt{InternVL3-1B}, which punch above their parameter scale, often outperforming heavier LLMs in overall accuracy. Notably, some high-parameter LLMs such as \texttt{Janus-Pro-7B} and \texttt{LLaVA-1.6-mistral-7B} exhibit fluctuating performances, suggesting that parameter size alone is not a sufficient predictor of effectiveness, especially in multilingual and multimodal tasks. At the lower end, reasoning-centric models like \texttt{Kimi-VL-A3B-Thinking} and less-tuned Indic LMs like \texttt{Chitrarth-1} show limited generalization, with accuracies significantly trailing in both zero-shot and CoT settings. The overall findings emphasize that well-aligned cross-modal reasoning and cultural grounding can outperform sheer scale in diverse evaluation settings.

\subsection{RQ2: How do vision-language models vary in performance across Indian languages with unequal resource support?}

\textbf{Answering RQ2: Language-wise Difficulty Spectrum}
A breakdown by language shows a clear gap between high- and low-resource contexts. English remains the most reliably understood language, as expected, with near-saturation accuracy levels for many models. This is followed by Hindi, Bengali, and Marathi, likely benefiting from better multilingual training corpora and shared Indo-Aryan linguistic roots. Conversely, languages like Sindhi, Konkani, and Kannada consistently pose the greatest challenges, with accuracy dropping by over 40\% in some cases compared to English. These disparities underscore systemic gaps in training data and cultural alignment in current VLMs. Moreover, languages like Assamese and Odia, despite their wide speaker base, do not exhibit uniformly high performance, hinting at underrepresentation in foundational model pretraining datasets. This highlights the urgent need for better linguistic inclusion, particularly for Indian languages at the tail-end of the accuracy distribution. A more detailed breakdown of state-wise and Union Territory–wise language performance accuracies is provided in Figure~\ref{fig:accuracy_zero_shot_models-state-wise} in the Appendix, due to space constraints.

\subsection{RQ3: What types of questions pose difficulties to current vision-language models?}

\textbf{Answering RQ3: Question Type-Specific Trends}
When segmented by question type, it becomes evident that \texttt{General Questions} and \texttt{Common Sense Cultural Questions} receive the highest accuracy across models, suggesting that these models are relatively proficient at surface-level understanding and culturally grounded inferences. However, \texttt{Multi-hop Reasoning Questions} introduce a steep drop in accuracy, exposing models' limitations in sequential inferencing and logical chaining. While CoT prompting helps moderately in lifting scores for this category, its gains are not uniformly robust across all languages. Additionally, \texttt{Analogy Questions} show the highest variance, some models excel when semantic similarity is explicit, while others flounder, reflecting a fragile grasp of abstract reasoning. These findings call for further attention toward reasoning scaffolds and prompt design that specifically target relational and inferential understanding.

\subsection{RQ4: How does model typology influence performance across task categories and languages?}

\textbf{Answering RQ4: Insights by Model Category}
Stratifying performance based on model typology yields several revealing patterns. \textbf{SLMs} such as \texttt{SmolVLM-256M-Instruct} and \texttt{InternVL3-1B} perform surprisingly well given their compact size, particularly excelling in general question - answering and commonsense tasks. \textbf{LLMs}, while expectedly powerful, do not always justify their computational footprint—models like \texttt{Qwen2-VL-7B} and \texttt{Llama-4-Scout-17B} show decent multilingual adaptability, but their gains plateau in deeper reasoning tasks. \texttt{Maya} demonstrates robust and balanced performance across multiple settings, outperforming several larger general-purpose LLMs in culturally grounded understanding. In contrast, other \textbf{Indic LMs}, such as \texttt{Chitrarth-1}, show comparatively weaker results, highlighting ongoing challenges in region-specific fine-tuning and alignment with image-grounded reasoning. Furthermore, \textbf{Reasoning-oriented models} like \texttt{Kimi-VL-A3B-Thinking} show promise in isolated tasks but fail to generalize across linguistic and logical variation. Finally, \textbf{Proprietary models} like \texttt{GPT-4o mini} remain the gold standard, consistently delivering the best zero-shot and CoT results across languages and question types, illustrating the strength of multi-modal scaling and integrated training pipelines. These insights collectively reinforce the need for balanced development across efficiency, reasoning, and multilingual inclusiveness. 

\subsection{RQ5: How does model performance differ between zero-shot and Chain-of-Thought (CoT) prompting across various question types, and which models benefit most from reasoning scaffolds?}

\textbf{Answering RQ5: Zero-shot vs. Chain-of-Thought (CoT) Performance Analysis}
We compare model performance under zero-shot and chain-of-thought (CoT) prompting to assess the value of explicit reasoning scaffolds. CoT proved most beneficial for reasoning-intensive categories such as multi-hop and analogy questions, yielding accuracy gains of up to 10--15\%, while common-sense cultural questions showed only modest improvements. Large-scale proprietary models (e.g., GPT-4o mini) consistently benefited across question types, whereas smaller instruction-tuned models (e.g., SmolVLM-256M-Instruct, InternVL3-1B) showed competitive gains, sometimes being on par with larger open-source systems. By contrast, reasoning-specialized (e.g., Kimi-VL-A3B-Thinking) and Indic-focused models (e.g., Chitrarth) exhibited limited or inconsistent improvements, suggesting weaker generalization of CoT in low-resource or culturally specific settings. Although CoT narrowed performance gaps on complex tasks, challenges in analogical reasoning and disparities across languages remain, with high-resource languages (e.g., Hindi, Bengali) benefiting more than low-resource ones (e.g., Konkani, Sindhi). Overall, CoT enhances culturally grounded reasoning, but its impact varies by question type, model family, and linguistic coverage. \emph{Due to limited space, we include the \textbf{error analysis} in the appendix \ref{error}.}

\section{Conclusion}
In this study, we introduced the DRISHTIKON dataset to evaluate the capabilities of vision-language models (VLMs) in the Indian cultural context. Spanning 15 diverse Indian languages, our evaluation across a range of VLMs uncovers several key insights. Proprietary models such as \texttt{GPT-4o mini} demonstrate strong performance, benefiting from large-scale instruction tuning and alignment. Notably, compact instruction-tuned models like \texttt{SmolVLM-256M-Instruct} and \texttt{InternVL3-1B} consistently deliver competitive results, highlighting the promise of efficiency-aware architectures for culturally rich tasks. Encouragingly, the Indian-origin \texttt{Maya} model also performed well, underscoring the potential of indigenous efforts in building culturally aligned and linguistically inclusive AI systems. Persistent performance gaps highlight digital inequities, with low-resource Indian languages trailing due to limited data and exposure. DRISHTIKON underscores the need for inclusive, culturally aware, and efficient VLMs, offering a robust benchmark for future multilingual research.

\section*{Limitations}
While the DRISHTIKON benchmark makes a significant step toward evaluating cultural and linguistic reasoning in Indian contexts, certain limitations remain. Despite covering 15 languages and diverse cultural settings, the dataset cannot exhaustively represent the full spectrum of India's regional nuances, dialectical variations, and socio-cultural practices. Additionally, although curated image-text pairs enable controlled evaluation, they may carry subtle annotator biases and may not fully replicate the complexity of real-world multimodal scenarios.

On the modelling side, even with the aid of Chain-of-Thought prompting, many VLMs continue to struggle with tasks involving abstract analogies and multi-hop reasoning, indicating room for improvement in compositional understanding. Furthermore, performance gaps across languages reflect the broader challenge of digital disparity, particularly for low-resource languages with limited training data. These insights highlight opportunities for future work in developing more inclusive datasets, culturally attuned training strategies, and robust reasoning frameworks that can support equitable and generalizable multimodal AI.

\section*{Ethics Statement}

\textbf{Data Sourcing and Cultural Integrity:}
The DRISHTIKON dataset was constructed using publicly available resources and licensed materials, ensuring adherence to data-sharing norms and copyright considerations. Care was taken to represent diverse linguistic and cultural contexts across India, with a focus on including both high- and low-resource languages. While every effort was made to maintain balance and inclusivity, we acknowledge that certain regional or dialectal variations may still be underrepresented due to the limitations of available data.

\textbf{Human Annotation and Fair Compensation:}
To ensure the cultural validity and linguistic accuracy of the dataset, we employed a team of annotators proficient in different Indian languages and familiar with their respective cultural contexts. Annotators were fairly compensated at an average hourly rate (in USD), and a detailed breakdown is included in the appendix. Training and guidelines were provided to mitigate personal or regional biases, and a validation step was conducted to ensure annotation consistency and cultural sensitivity. Efforts were made to avoid harmful stereotypes and to ensure questions reflect respectful and inclusive representations.

\textbf{Responsible Use and Community Benefit:}
DRISHTIKON was developed with the intention to support the development of culturally aware, multilingual vision-language models. We encourage its use in academic and research settings that promote fairness, inclusivity, and transparency in AI. Any misuse of the dataset for generating biased, discriminatory, or culturally insensitive outputs would go against the values and intent behind its creation.

\textbf{Licensing and Permissible Use:}
The DRISHTIKON dataset is released strictly for research and non-commercial use. To avoid copyright infringement, we provide only URLs pointing to publicly available images rather than hosting the images directly. These URLs are intended to be used for academic reference, ensuring compliance with fair use principles and image-sharing policies. Users of the dataset are expected to respect the original source licenses and terms of use when accessing or displaying these images.

\section*{Acknowledgments}
Akash Ghosh and Sriparna Saha express gratitude to the SERB POWER scheme (SPG/2021/003801), Department of Science and Technology, Government of India.

\bibliography{custom}

\appendix

\section{Appendix}

\subsection{Data Sources and Selection Criteria}
\label{app-1}

The construction of the \textit{\textbf{DRISHTIKON}} benchmark involved a rigorous and multi-phased data curation process to ensure a balanced, authentic, and representative coverage of India's diverse socio-cultural fabric. The following publicly accessible and reputed platforms were employed as primary sources of information:

\begin{itemize}
    \item \textbf{Wikipedia}\footnote{\url{https://www.wikipedia.org}}: Served as a foundational source offering encyclopedic, well-referenced summaries on Indian festivals, attire, regional cuisines, monuments, and personalities. Special care was taken to cross-check citations for factual accuracy.
    
    \item \textbf{Ritiriwaz}\footnote{\url{https://www.ritiriwaz.com}}: A culturally focused platform that provided in-depth articles on Indian customs, rituals, marriage traditions, and ethnic wear, capturing nuances often absent from generic encyclopedic sources.
    
    \item \textbf{Holidify}\footnote{\url{https://www.holidify.com}}: Primarily used for region-specific insights, including local attractions, cultural highlights, state-wise festivals, and seasonal events, aiding in geographically diverse content gathering.
    
    \item \textbf{Google Arts \& Culture}\footnote{\url{https://artsandculture.google.com}}: Offered high-quality curated exhibits on Indian art, dance forms, textiles, and heritage monuments, with visual and narrative depth suitable for grounding vision-language tasks.
    
    \item \textbf{Times of India}\footnote{\url{https://timesofindia.indiatimes.com}}: A leading news platform that supplemented static knowledge with contemporary coverage of cultural events, notable figures, and evolving regional practices.
\end{itemize}

These sources were chosen for their complementary strengths—ranging from encyclopedic objectivity and regional specificity to cultural richness and visual storytelling. Selection was guided by criteria such as factual reliability, diversity of representation across Indian states and domains, granularity of cultural context, and availability of multimodal content. Redundancy was minimized by cross-referencing facts, and only those entries substantiated by multiple sources were retained for MCQ generation. This curated corpus underpins the evaluation benchmark and ensures that generated questions holistically reflect India’s heterogeneous cultural identity.

\begin{figure}[h]
\centering
\includegraphics[width=0.45\textwidth]{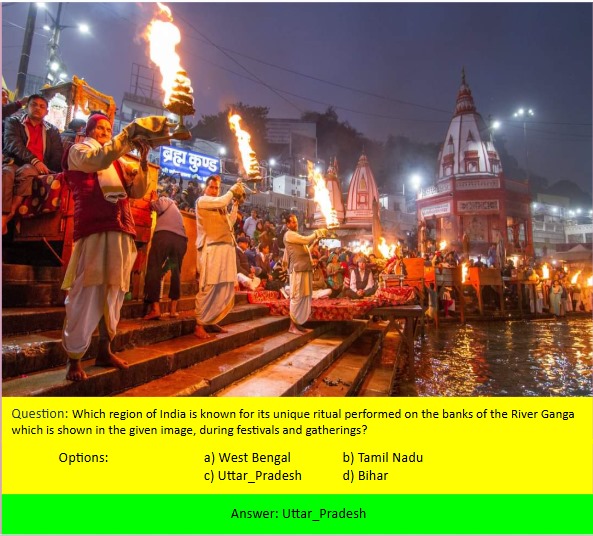}
\caption{Example of a visual MCQ associated with the River Ganga Aarti in Uttar Pradesh.}
\label{fig:mcq-image}
\end{figure}

\subsection{Annotation Methodology and Validation Protocol}
\label{app-2}

To ensure the quality and cultural fidelity of the \textit{\textbf{DRISHTIKON}} benchmark, we implemented a multi-stage validation process for the 2,126 multimodal multiple-choice questions (MCQs), as shown in an example in Figure~\ref{fig:mcq-image}.

\subsubsection{Annotation Workflow}
Each MCQ was initially authored by trained annotators with backgrounds in Indian history, sociology, or cultural studies. Annotators followed structured guidelines that emphasized:

\begin{itemize}
    \item \textbf{Cultural authenticity:} Questions were crafted to reflect regionally grounded knowledge and practices, avoiding stereotypes or generic generalizations.
    \item \textbf{Clarity and neutrality:} Question stems and options were phrased in clear, neutral language, avoiding suggestive cues or complex phrasing that could bias responses.
\end{itemize}

\subsubsection{Validation Process}
We employed a two-pass validation process:
\begin{enumerate}
    \item \textbf{Pass 1 – Peer Review:} Each question was independently reviewed by another annotator for factual accuracy, linguistic clarity, and option plausibility. Any ambiguities or factual discrepancies were flagged and corrected.
    \item \textbf{Pass 2 – Expert Adjudication:} A cultural expert with domain knowledge performed a final adjudication step to resolve edge cases and confirm correctness.
\end{enumerate}

\subsubsection{Agreement and Quality Control}
To assess consistency, we calculated inter-annotator agreement (IAA) on a random 20\% sample of the MCQs using Cohen’s $\kappa$. We observed substantial agreement ($\kappa = 0.82$) between initial annotators and peer reviewers. Disagreements primarily arose from regional overlaps (e.g., shared traditions across bordering states), which were resolved through discussion or expert input.

\subsubsection{Visuals Check}
For visual context, annotators referenced image metadata and cross-verified content against at least two textual sources. Visuals were validated with the same two-pass process and were checked to ensure the image did not overtly reveal the answer through text overlays or location tags.

This meticulous pipeline ensured that the benchmark questions are reliable, culturally inclusive, and suitable for robust multimodal evaluation.

\subsection{Cultural Taxonomy and Attribute Definitions}
\label{app-4}

To support structured cultural benchmarking in the \textbf{DRISHTIKON} dataset, each question-image pair was tagged with a cultural attribute. The attributes were drawn from a dynamic taxonomy that reflects the breadth and complexity of India's socio-cultural heritage. Below are the attribute categories and their working definitions:

\begin{itemize}
    \item \textbf{Art:} Visual and decorative arts including painting, sculpture, traditional crafts, and region-specific artistic practices.

    \item \textbf{Costume:} Traditional attire, region-specific garments, and symbolic clothing worn during rituals, festivals, or daily life.

    \item \textbf{Cuisine:} Food items, cooking practices, regional dishes, and culinary customs that characterize Indian states or communities.

    \item \textbf{Cultural Common Sense:} Widely known cultural facts, idioms, practices, or behaviors that are intuitive to locals but may not be explicitly taught.

    \item \textbf{Dance and Music:} Classical, folk, and contemporary forms of dance and music tied to regional or religious traditions.

    \item \textbf{Festivals:} Celebrations, fairs, and religious or seasonal festivals observed across different Indian regions and communities.

    \item \textbf{History:} Historical figures, events, timelines, or periods that shaped India’s regional and national identity.

    \item \textbf{Language:} Native languages, dialects, scripts, and linguistic practices across different states and territories.

    \item \textbf{Medicine:} Traditional healing systems such as Ayurveda, Siddha, Unani, and folk medical practices and their cultural relevance.

    \item \textbf{Nightlife:} Cultural expressions of nightlife including entertainment, food, rituals, and urban evening practices specific to regions.

    \item \textbf{Personalities:} Notable figures in culture, politics, arts, science, or social reform with significant cultural influence.

    \item \textbf{Religion:} Religious symbols, rituals, deities, and practices across India's major and minor religious communities.

    \item \textbf{Rituals and Ceremonies:} Practices associated with worship, rites of passage, or daily cultural-religious observances.

    \item \textbf{Sports:} Traditional and modern sports, indigenous games, and regionally popular athletic events or personalities.

    \item \textbf{Tourism:} Destinations, experiences, or features that are central to domestic or international tourism in India.

    \item \textbf{Transport:} Culturally symbolic or region-specific modes of transport including boats, bullock carts, local trains, and more.
\end{itemize}

\subsubsection{Attribute Tagging Methodology}

To support a culturally-aware evaluation of models, each multiple-choice question (MCQ) in our dataset was manually tagged with a single cultural attribute by trained annotators. The attributes span categories such as \texttt{Festivals}, \texttt{Rituals\_and\_Traditions}, \texttt{Attire}, \texttt{Art\_Forms}, \texttt{Language}, \texttt{Cuisine}, \texttt{Geography}, and \texttt{Historical\_Heritage}, among others.

Figure~\ref{fig:mcq-image} illustrates an example where a seemingly ambiguous question could potentially fall under multiple cultural categories. The image-based MCQ features the Ganga Aarti performed on the banks of the River Ganga in Varanasi. While one might consider tagging it under \texttt{Festivals} due to its grand and ceremonial appearance, our annotation guidelines emphasized tagging based on the most representative and contextually consistent interpretation.

\begin{figure*}[htbp]
\centering
\includegraphics[width=\textwidth]{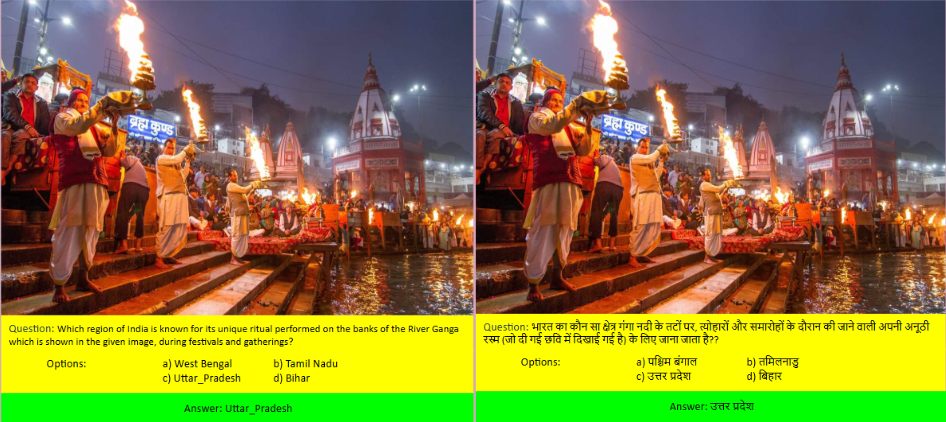}
\caption{Example of MCQ translated from English to Hindi.}
\label{fig:translation_comparison}
\end{figure*}

In this case, the question was tagged under \texttt{Rituals\_and\_Traditions} because the Ganga Aarti is not confined to a specific festival—it is a daily ritual deeply embedded in local tradition and spiritual practice. Such nuanced decisions were made through annotator deliberation and cross-verification to ensure clarity and precision in tag assignment.

Ambiguities—such as overlap between festive and ritualistic cues—were discussed and adjudicated collectively. Only one attribute was assigned per MCQ to facilitate clean categorization and dataset slicing for downstream evaluation tasks.

This tagging strategy ensures that even culturally complex instances are consistently annotated, allowing researchers to probe model performance across diverse yet unambiguous cultural dimensions.

\subsection{Sampling and Reasoning-based Augmentation}
\label{app-5}

To ensure balanced evaluation across geographic regions, we introduced a reasoning-based augmentation phase using a stratified subset of 720 MCQs (20 per region from 36 Indian states and union territories). This uniform count was guided by the region with the lowest question availability, thereby avoiding data imbalance during augmentation.

\subsubsection{Stratified Sampling from Richer Regions}
For regions that originally had more than 20 MCQs, we employed a stratified sampling approach grounded in attribute coverage. Each MCQ in our dataset was previously tagged with one of several cultural attributes. When selecting the subset of 20 questions for such regions, we ensured that this attribute distribution remained approximately proportional to that in the full regional set.

For instance, if the state of West Bengal had 60 MCQs—20 focused on \texttt{Festivals}, 15 on \texttt{Cuisine}, 10 on \texttt{Attire}, and 15 on \texttt{Historical\_Heritage}—then the selected 20-question subset maintained this diversity using proportional sampling:
\begin{itemize}
    \item 7 questions from \texttt{Festivals}
    \item 5 from \texttt{Cuisine}
    \item 3 from \texttt{Attire}
    \item 5 from \texttt{Historical\_Heritage}
\end{itemize}

In cases where exact proportionality was not feasible due to rounding or attribute sparsity, we prioritized inclusion of underrepresented cultural aspects to ensure thematic balance. This approach not only preserved intra-regional diversity but also prevented dominance of popular attributes (e.g., \texttt{Festivals}) in regions with rich cultural repositories.

\begin{figure*}[htbp]
\centering
\includegraphics[width=\textwidth]{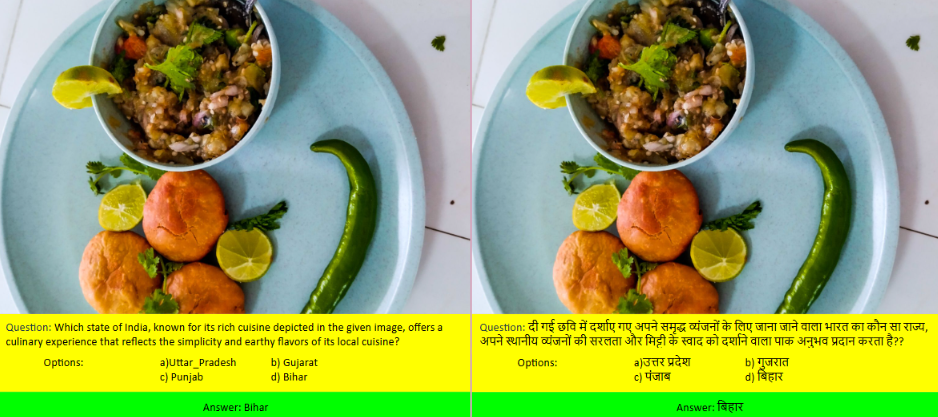}
\caption{Another example of MCQ translated from English to Hindi.}
\label{fig:translation_comparison2}
\end{figure*}

\subsubsection{Why Stratified Sampling?}
Simple random sampling could have led to subsets skewed toward the most frequent attribute in that region (e.g., \texttt{Festivals}), thereby reducing cultural variety. Our stratified method guaranteed that rare but significant cultural dimensions (like \texttt{Performing\_Arts} or \texttt{Attire}) were also retained in the reasoning-based augmented set.

This strategic curation enhances the fairness and comprehensiveness of model evaluations, enabling consistent benchmarking of cultural understanding across both high-resource and low-resource regions.

\subsubsection{Validation}
Subset composition was validated post hoc through a comparison of attribute distributions before and after sampling. Pearson's chi-squared tests showed no statistically significant loss in attribute variety (p > 0.1), affirming that the sampling retained cultural diversity within a manageable subset size.

\subsection{Translation Quality and Human Verification Protocol}
\label{app-6}

To evaluate the quality of translations across culturally rich and semantically nuanced questions, we present two illustrative examples comparing the English source to their Hindi translations. Figure~\ref{fig:translation_comparison} and~\ref{fig:translation_comparison2}  showcase translated samples—one referencing a ritual (Ganga Aarti) and the other a regional dish (Litti Chokha)—used to assess our multilingual pipeline.

We utilized the Gemini Pro~\cite{deepmind2025geminipro} language model for translation, motivated by its strong multilingual semantic fidelity and contextual grounding, as demonstrated on FLORES-200 and XTREME-UP benchmarks~\cite{nllb2022,ref1,ref2}. Its ability to handle idiomatic and domain-specific expressions made it suitable for our linguistically and culturally diverse dataset.

\paragraph{Human Verification Protocol.}
To mitigate risks of hallucination or mistranslation\cite{sahoo2024unveiling}, a two-stage human verification pipeline was adopted:
\begin{itemize}
    \item \textbf{Stage 1:} Bilingual reviewers verified semantic consistency, fluency, and adherence to the original question’s intent on stratified samples.
    \item \textbf{Stage 2:} A separate round of quality control ensured inter-annotator agreement and cultural appropriateness.
\end{itemize}

\paragraph{Evaluation of Translations.}
In both examples:
\begin{itemize}
    \item \textbf{Semantic fidelity} is preserved. For instance, in the first example (Figure~\ref{fig:translation_comparison}), the phrase ``ritual performed on the banks of the River Ganga” is translated into Hindi with appropriate syntactic structure and vocabulary, keeping the reverent tone intact.
    \item \textbf{Cultural relevance} is maintained. In the second example (Figure~\ref{fig:translation_comparison2}), describing the cuisine of Bihar, the translation preserves key descriptors to retain the earthy connotation associated with ``simplicity and earthy flavours.”
\end{itemize}

\begin{table*}[htbp]
\centering
\caption{Number of Annotators per Language/State and Average Pay}
\begin{tabular}{|l|c|c|}
\hline
\textbf{Language / State} & \textbf{No. of Annotators} & \textbf{Avg. Pay per Hour (USD)} \\
\hline
Hindi (Uttar Pradesh, Bihar)        & 10 & 3.00 \\
Bengali (West Bengal)               & 6  & 2.88 \\
Tamil (Tamil Nadu)                  & 5  & 3.12 \\
Telugu (Andhra Pradesh, Telangana)  & 8  & 3.00 \\
Kannada (Karnataka)                 & 4  & 2.88 \\
Malayalam (Kerala)                 & 3  & 3.00 \\
Marathi (Maharashtra)              & 6  & 2.94 \\
Gujarati (Gujarat)                  & 4  & 2.88 \\
Punjabi (Punjab)                    & 3  & 2.82 \\
Assamese (Assam)                    & 3  & 2.76 \\
Odia (Odisha)                       & 3  & 2.76 \\
Urdu (Delhi, Jammu \& Kashmir)      & 4  & 3.00 \\
Others (e.g., Sikkim, Ladakh)       & 2  & 2.64 \\
\hline
\end{tabular}
\label{tab:annotator-stats}
\end{table*}

\paragraph{Challenges and Resolutions.}
\begin{itemize}
    \item \textbf{Syntactic divergence}: Hindi sentence structures often require reordering of clauses. For instance, direct translations can result in unnatural phrasing. We prompted Gemini Pro to produce natural, idiomatic Hindi and post-edited awkward constructs.
    \item \textbf{Cultural terminology}: Some terms (e.g., ``Aarti'' or ``Litti Chokha'') lack equivalents. We opted for \emph{transliteration} or descriptive phrases when appropriate, preserving cultural identity while ensuring comprehension.
    \item \textbf{Lexical alignment}: Ambiguities in English adjectives like ``rich” or ``earthy” were contextually resolved using local equivalents in Hindi, guided by cultural connotation rather than direct word-to-word substitution.
\end{itemize}

Overall, this semi-automated translation + verification workflow allowed us to scale high-quality multilingual data curation while maintaining semantic, syntactic, and cultural integrity.

\begin{figure}[htbp]
  \centering
  \includegraphics[width=0.45\textwidth]{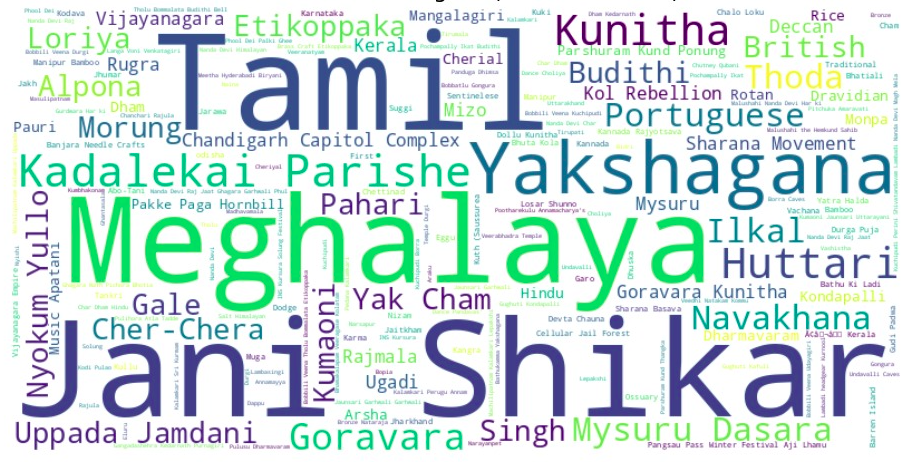}
  \caption{Culturally-Specific Vocabulary in \textsc{Drishtikon}}
  \label{fig:cultural-wordcloud}
\end{figure}

\subsection{Annotator Details by Language/State}\label{annotators}
We employed human annotators from diverse Indian states and linguistic backgrounds to ensure cultural sensitivity, regional nuance, and language-specific accuracy in the annotation process (shown in Table \ref{tab:annotator-stats}). The selection aimed to balance representation across both high-resource and low-resource languages. Annotators were recruited based on their fluency in the respective regional languages and their educational qualifications (minimum: bachelor's degree). Prior to the task, all annotators underwent training sessions to familiarize themselves with the guidelines and quality expectations. Compensation was provided on an hourly basis, reflecting fair labour standards and encouraging consistent performance. The table below summarizes the number of annotators per language or state, along with the average hourly pay (in USD).

\subsection{Further Dataset Statistics}
\label{app-8}

To offer deeper insight into the structure and cultural span of the \textsc{Drishtikon} dataset, we provide extended statistical breakdowns through visualizations and metadata summaries.

\subsubsection{Word Cloud Analyses}

\begin{figure}[htbp]
  \centering
  \includegraphics[width=0.45\textwidth]{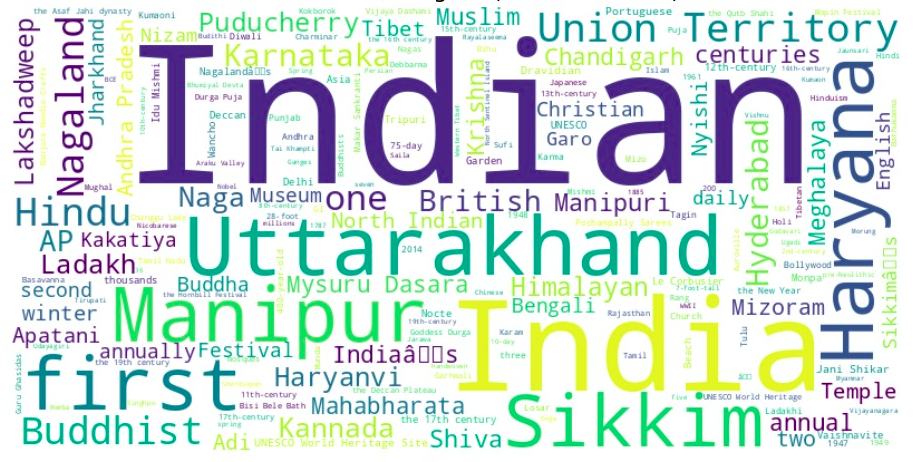}
  \caption{Full Vocabulary Distribution Across All Questions}
  \label{fig:full-question-wordcloud}
\end{figure}

\begin{figure*}[htbp]
    \centering
    \includegraphics[width=\textwidth]{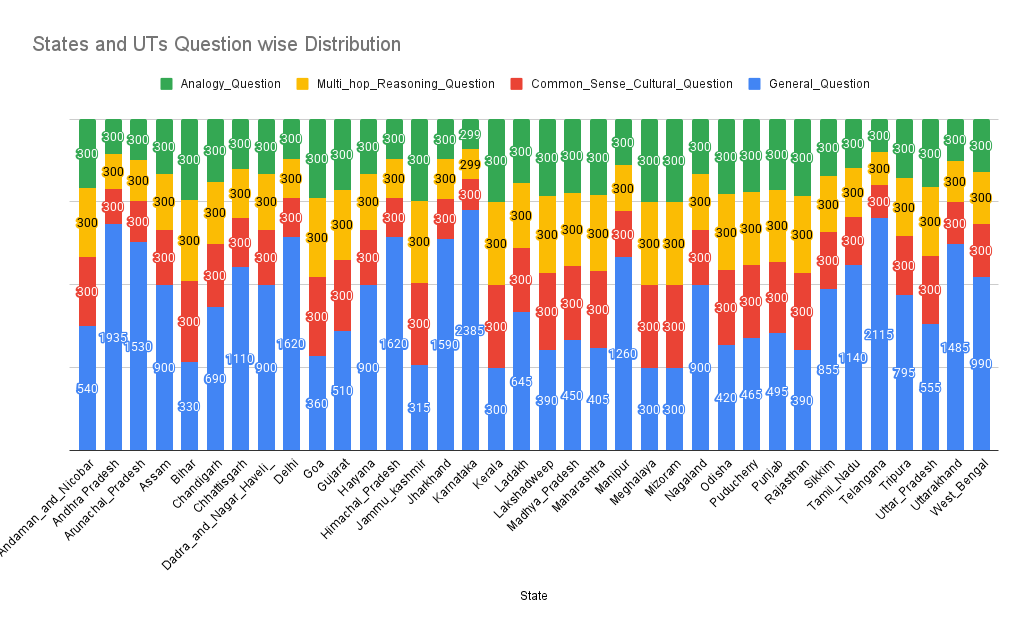}
    \caption{State-wise distribution of question types across Indian regions.}
    \label{fig:state-wise-distribution}
\end{figure*}

We visualize the most salient terms for English component of our dataset using two complementary word clouds:

\begin{itemize}
    \item \textbf{Culturally-Specific Elements:} This word cloud (Figure~\ref{fig:cultural-wordcloud}) captures culturally grounded concepts, traditions, festivals, and regionally rooted lexicon sourced from our question stems and options. Prominent terms like \textit{Jani Shikar}, \textit{Yakshagana}, \textit{Meghalaya}, \textit{Tamil}, and \textit{Mysuru Dasara} suggest that the dataset richly represents diverse socio-cultural phenomena.

    \item \textbf{Full Question Corpus Vocabulary:} A second word cloud (Figure~\ref{fig:full-question-wordcloud}) is generated over the complete corpus of questions. It reflects broader linguistic themes and signals topical diversity. Frequent mentions of \textit{India}, \textit{first}, \textit{Haryana}, \textit{Manipur}, and \textit{Union Territory} indicate a strong presence of both general knowledge and region-specific focus.
\end{itemize}

\subsubsection{State-wise Question Distribution.}
Figure~\ref{fig:state-wise-distribution} illustrates the distribution of questions across all 36 Indian states and union territories, categorized into four types: General Questions, Common Sense Cultural Questions, Multi-hop Reasoning Questions, and Analogy Questions. While every region was balanced with 300 questions per category for the latter three types, the number of General Questions varied significantly across regions, reflecting data availability and population-specific cultural variance. This visualization underscores the heterogeneity in question volume, motivating our uniform 20-question sampling strategy for reasoning-based augmentation.

\subsection{Model Hyperparameter Settings}
\label{app:hyperparams}

The detailed hyperparameter settings used in our experiments are summarized in Table \ref{tab:hyperparams-table} for reference.

\begin{table*}[htbp]
\centering
\small
\begin{tabular}{lcccc}
\toprule
\textbf{Model} & \textbf{Size (Params)} & \textbf{Vision Encoder} & \textbf{Image Res.} & \textbf{Max Tokens} \\
\midrule
SmolVLM-256M-Instruct & 256M & ViT-B/16 & $224\times224$ & 1024 \\
InternVL3-1B & 1B & InternImage-L & $448\times448$ & 2048 \\
Janus-Pro-7B & 7B & CLIP-style & $336\times336$ & 4096 \\
Qwen2-VL-7B-Instruct & 7B & ViT-G & $448\times448$ & 8192 \\
LLaVA-1.6-Mistral-7B & 7B & CLIP-L/14 & $336\times336$ & 4096 \\
InternVL3-14B & 14B & InternImage-H & $448\times448$ & 4096 \\
Llama-4-Scout-17B & 17B & CLIP-style & $336\times336$ & 8192 \\
Gemma-3-27B-IT & 27B & Unknown & $224\times224$ & 8192 \\
Qwen2.5-Omni-7B & 7B & ViT-style & $448\times448$ & 8192 \\
\midrule
Kimi-VL-A3B-Thinking & 3B & ViT (proprietary) & $336\times336$ & 8192 \\
GPT-4o & - & Proprietary & $~512\times512$ & 128k (context) \\
\midrule
Chitrarth & - & Unknown & $224\times224$ & Unknown \\
Maya & 7B & CLIP-L/14 & $224\times224$ & 4096 \\
\bottomrule
\end{tabular}
\caption{Summary of hyperparameters for evaluated vision-language models. Where official details are unavailable, publicly documented defaults or best estimates are provided.}
\label{tab:hyperparams-table}
\end{table*}

\subsection{Error Analysis}\label{error}

While GPT-4o-mini demonstrated consistently strong performance across multilingual QA tasks, it occasionally produced incorrect answers. To gain deeper insights into these instances, we conducted a manual analysis of selected failure cases, a few of which are illustrated below.

Each example comprises the original English question, its associated image, and the model's incorrect prediction. These cases shed light on nuanced challenges that persist even for advanced language models.

Our analysis suggests that the observed errors stemmed from:
\begin{itemize}
    \item \textbf{Fine-grained semantic confusion} — particularly when distractor options were semantically close to the correct answer.
    \item \textbf{Over-reliance on lexical cues} rather than a comprehensive understanding of the context, especially in culturally nuanced questions.
    \item \textbf{Gaps in visual grounding} where accurate interpretation required deeper regional or cultural knowledge.
\end{itemize}

The examples discussed below are accompanied by interpretive commentary, highlighting opportunities to further enhance the multimodal and multilingual reasoning capabilities of such models.

\subsubsection{Error Case 1: Historical Leader Identification}

\begin{figure}[htbp]
    \centering
    \includegraphics[width=0.5\textwidth]{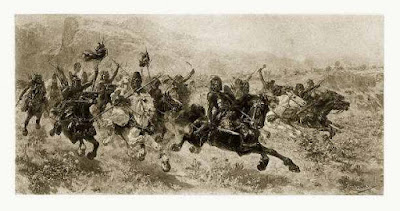}
    \caption{Depiction of a tribal uprising on horseback}
    \label{fig:kol_rebellion}
\end{figure}

\textbf{Question:} Who was the prominent leader of the depicted Rebellion? (Associated image: Figure \ref{fig:kol_rebellion})

\noindent\textbf{Options:}
\begin{enumerate}
    \item Budhu Bhagat
    \item Tilka Manjhi
    \item Sidho and Kanho Murmu
    \item Birsa Munda
\end{enumerate}

\noindent\textbf{Model Output:} Option 3 = \textit{Sidho and Kanho Murmu} \\
\textbf{Correct Answer:} \textit{Budhu Bhagat}

\noindent\textbf{Error Intuition:}  
The model likely associated the visual of tribal warriors on horseback with the more widely recognized Santhal Rebellion led by Sidho and Kanho Murmu, rather than the Kol Rebellion led by Budhu Bhagat. Given that both rebellions share thematic similarities—tribal resistance, traditional attire, and armed revolt—the model appears to have relied on surface-level visual patterns and the popularity of certain leaders, rather than grounding the answer in historical specificity or regional cues.

\subsubsection{Error Case 2: Misclassification of Cultural Dance Form}

\begin{figure}[htbp]
    \centering
    \includegraphics[width=0.45\textwidth]{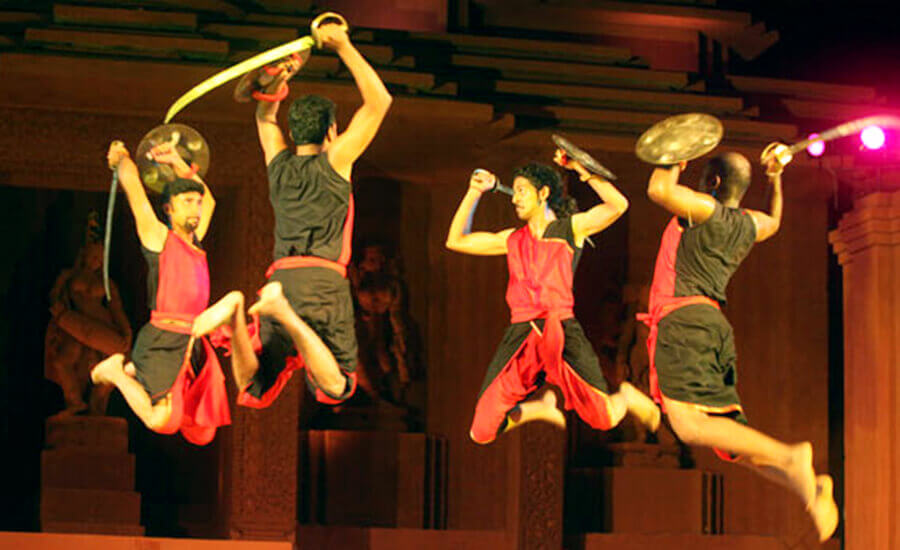}
    \caption{Depiction of a traditional martial dance performance}
    \label{fig:paika_dance}
\end{figure}

\textbf{Question:} The depicted dance, a unique art form blending martial arts with rhythmic movements and performed exclusively by men, originates from which Indian state?  (Associated image: Figure \ref{fig:paika_dance})

\noindent\textbf{Options:}
\begin{enumerate}
    \item Chhattisgarh
    \item Jharkhand
    \item West Bengal
    \item Odisha
\end{enumerate}

\noindent\textbf{Model Output:} Option 4 = \textit{Odisha} \\
\textbf{Correct Answer:} \textit{Jharkhand}

\noindent\textbf{Error Intuition:}  
The model incorrectly predicted Odisha, possibly confusing the dance with the similarly martial-themed ``Paika” dance of Odisha, which also involves weapons and is visually comparable. The correct answer, however, is the ``Paika Akhara" of Jharkhand. This confusion likely stems from visual and thematic overlap between regional martial dances, and the model’s bias toward more widely documented or referenced traditions in training data.

\subsubsection{Error Case 3: Misidentification of Tribal Art Form}

\begin{figure}[htbp]
    \centering
    \includegraphics[width=0.5\textwidth]{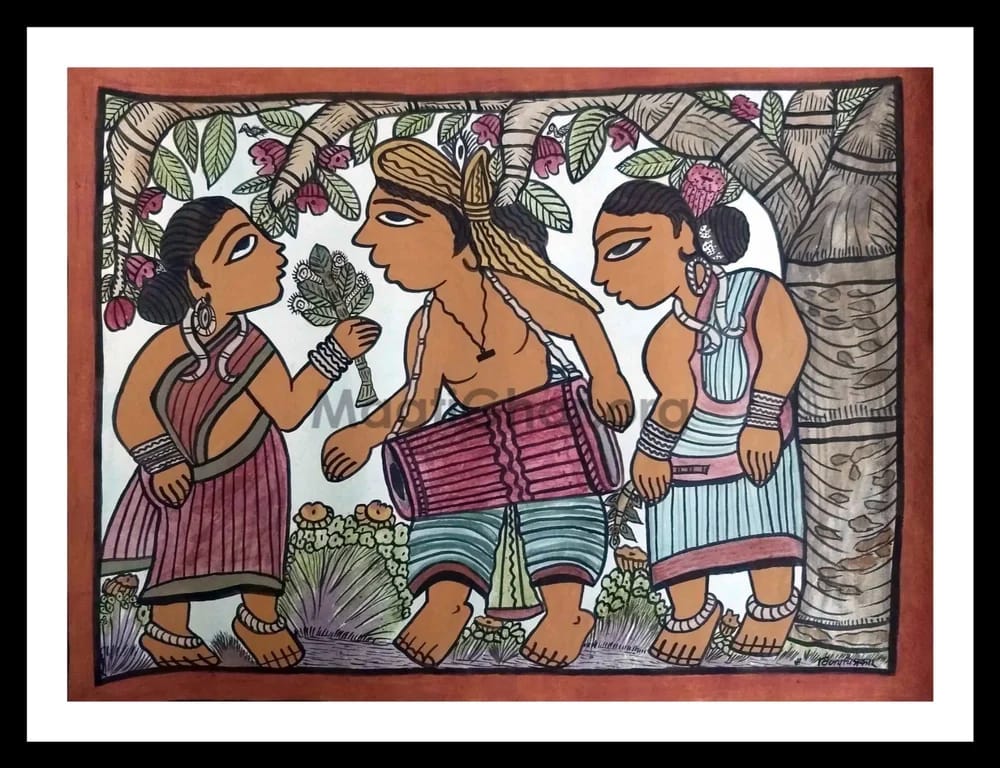}
    \caption{Paitkar painting – a traditional scroll painting style}
    \label{fig:paitkar_painting}
\end{figure}

\textbf{Question:} The paintings depicted in the image, one of the oldest tribal art forms in India, originated in which state? (Associated image: Figure \ref{fig:paitkar_painting})

\noindent\textbf{Options:}
\begin{enumerate}
    \item Jharkhand
    \item Tamil Nadu
    \item Punjab
    \item Gujarat
\end{enumerate}

\noindent\textbf{Model Output:} Option 4 = \textit{Gujarat} \\
\textbf{Correct Answer:} \textit{Jharkhand}

\noindent\textbf{Error Intuition:}  
The model incorrectly identified the origin as Gujarat, possibly confusing the Paitkar painting style with more globally recognized folk arts like Warli or Pithora. The correct answer is Jharkhand, where the Paitkar art form—believed to be one of India’s earliest scroll painting traditions—emerged. The misclassification likely stems from the model's underexposure to tribal art forms from eastern India in its pretraining data.

\subsubsection{Error Case 4: Misclassification of Cultural Landmark Location}

\begin{figure}[htbp]
    \centering
    \includegraphics[width=0.45\textwidth]{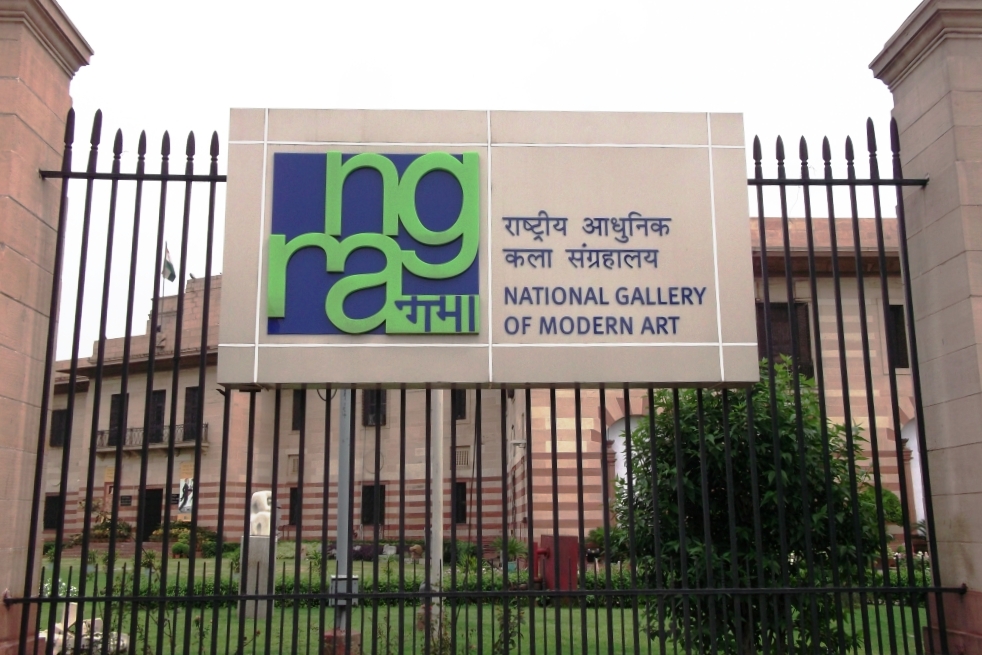}
    \caption{Signboard of the National Gallery of Modern Art}
    \label{fig:ngma_delhi}
\end{figure}

\textbf{Question:} The depicted Gallery is located in which city? (Associated image: Figure \ref{fig:ngma_delhi})

\noindent\textbf{Options:}
\begin{enumerate}
    \item Pune
    \item Mumbai
    \item Delhi
    \item Hyderabad
\end{enumerate}

\noindent\textbf{Model Output:} Option 2 = \textit{Mumbai} \\
\textbf{Correct Answer:} \textit{Delhi}

\noindent\textbf{Error Intuition:}  
Despite the clear signage in both English and Hindi indicating the National Gallery of Modern Art (NGMA), the model incorrectly associated it with Mumbai. This confusion likely stems from the presence of NGMA branches in Mumbai and Bengaluru; however, the headquarters and the most iconic building is in New Delhi. The model failed to distinguish the specific architecture and setting unique to the Delhi branch.

\subsection{COT prompt}
\label{cot}

Our prompt leverages a culturally grounded chain-of-thought reasoning framework inspired by classical Indian epistemology. It guides the model to analyze images and questions through four distinct dimensions—visual insight, cultural memory, logical integration, and regional contextualization—to arrive at accurate, culturally informed answers. The design encourages nuanced reasoning while ensuring concise output by restricting the response to the final correct option only.

\begin{figure*}[htbp]
\centering
\begin{tcolorbox}[colback=blue!5!white,
                  colframe=blue!60!black,
                  title=\textbf{Prompt 4: Cultural Chain-of-Thought Reasoning (Drishti–Smriti–Yukti–Sthiti)},
                  width=\textwidth,
                  boxrule=0.6pt,
                  arc=4pt,
                  left=4pt,
                  right=4pt,
                  top=4pt,
                  bottom=4pt,
                  coltitle=white]
\textbf{Role:} You are an expert analyst deeply knowledgeable in Indian culture, traditions, and regional heritage. Carefully analyze the provided image and question. Reason methodically through each of the following culturally informed dimensions to identify the correct answer. Please output only the correct option/answer from the given options without any additional information or reasoning steps.

\textbf{Dimension A – Drishti (Visual Insight):} Carefully examine the image, identifying culturally significant visual elements such as attire, architecture, rituals, landscapes, or symbols.

\textbf{Dimension B – Smriti (Cultural Memory):} Recall relevant historical details, traditional knowledge, or well-known cultural practices from India related to this question.

\textbf{Dimension C – Yukti (Logical Integration):} Logically integrate your observations from Drishti and knowledge from Smriti. Use this integration to rule out options that are culturally or logically inconsistent.

\textbf{Dimension D – Sthiti (Regional Contextualization):} Consider regional and cultural contexts within India. Determine which of the provided options best aligns with the cultural and regional insights you've gained.

\textbf{Justification:} This prompt introduces a culturally grounded chain-of-thought reasoning process, drawing from classical Indian epistemology. By structuring the reasoning into Drishti (seeing), Smriti (remembering), Yukti (reasoning), and Sthiti (situating), it elicits explainable and culturally nuanced inference. The final constraint—returning only the correct option—ensures concise evaluation of the model’s internalized reasoning capacity without relying on verbosity.

\end{tcolorbox}
\end{figure*}

\begin{figure*}[htbp]
\centering
\includegraphics[width=\textwidth]{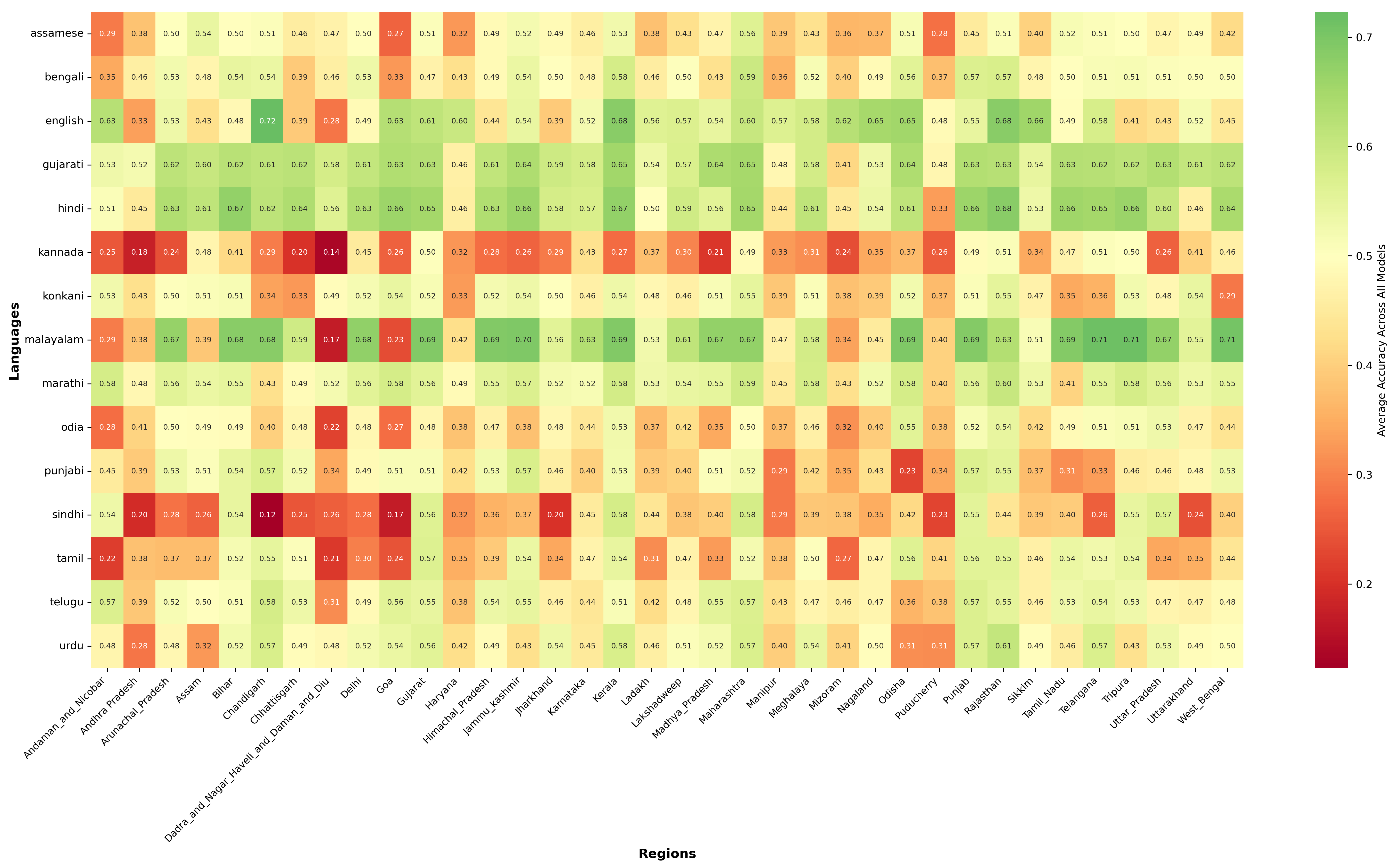}
\caption{State-wise and Union-territories-wise accuracy across languages, reflecting the overall multilingual performance of all the models across states and languages.}
\label{fig:accuracy_zero_shot_models-state-wise}
\end{figure*}

\subsection{Frequently Asked Questions (FAQs)}

\begin{itemize}
    
\item \textbf{Q1. What is the main goal of our study?} 
The primary goal is to evaluate the cultural reasoning capabilities of language models (LMs) through multimodal prompts that incorporate images of cultural artifacts and require contextual, symbolic, or multi-hop reasoning.

\item \textbf{Q2. Why is culture-specific question generation important?}
Generic QA benchmarks often overlook culturally grounded reasoning. Our prompts introduce challenges that simulate real-world, heritage-driven understanding—crucial for building globally inclusive AI systems.

\item \textbf{Q3. What role does the image play in our prompts?}
Images act as anchors for cultural artifacts or symbols. Prompts explicitly refer to these visuals (``as referenced in the image”) to encourage multimodal grounding in the model’s response.

\item \textbf{Q4. How does our Cultural Chain-of-Thought prompt differ from standard CoT?}
Our prompt is inspired by classical Indian epistemological constructs—Drishti (perception), Smriti (memory), Yukti (reason), and Sthiti (contextualization)—to guide LLMs in culturally coherent decision-making.

\item \textbf{Q5. Why use separate prompts for commonsense, multi-hop, and analogy?}
Each prompt targets a different cognitive skill—commonsense cultural reasoning, multi-step inference, and symbolic analogy—to provide a diverse and diagnostic evaluation of model understanding.

\item \textbf{Q6. Where can one find the actual prompts and examples?}
All prompt templates, justifications, and example outputs are included in the Appendix.

\item \textbf{Q7. How do we ensure fair comparison across models?}
All models were provided the same image-question pairings and prompts.

\end{itemize}

\subsection{Prompt Designs for Different Question Types}\label{Q-types}

In this section, we provide the prompt templates used to generate three different question types across our multilingual multimodal setup. Each prompt was carefully crafted to probe different cognitive dimensions—commonsense cultural grounding, multi-hop logical chaining, and analogical reasoning. Below, we describe each prompt, its justification, and include illustrative examples to clarify their operationalization.

\begin{figure*}[htbp]
\centering
\begin{tcolorbox}[colback=blue!5!white,colframe=blue!75!black,title=\textbf{Prompt 1: Cultural Commonsense Question}]
\textbf{Role:} You are a cultural expert. Given a simple factual question, generate a reasoning-based version that requires cultural commonsense to answer. Avoid directly naming the answer. Include contextual or narrative clues instead. It is necessary to add a reference to the image of the cultural artifact like \textit{``as referenced in the image”}.

\textbf{Justification:} This prompt evaluates the model’s ability to engage with culturally grounded knowledge that is not explicitly stated, testing its implicit reasoning capabilities and contextual understanding of visual cues tied to heritage or tradition.

\textbf{Example}\\
\textit{General Question:} Known for its crescent-shaped edge and association with Bengali kitchen traditions, what is the tool depicted in the image used primarily for cutting vegetables? \textit{(as referenced in the image)}\\
\textit{Answer:} Boti
\end{tcolorbox}
\end{figure*}

\begin{figure*}[htbp]
\centering
\begin{tcolorbox}[colback=orange!5!white,colframe=orange!75!black,title=\textbf{Prompt 2: Multi-hop Reasoning Question}]
\textbf{Role:} Transform the following factual question into a multi-hop reasoning question. The answer should require at least two connected facts to arrive at the final response. Add cultural or historical information to guide reasoning. It is necessary to add a reference to the image of the cultural artifact like \textit{``as referenced in the image”}. DO NOT include any prefixes or labels like `Transformed question:’. Return ONLY the rewritten question without any additional text.

\textbf{Justification:} This prompt probes the model's ability to connect multiple pieces of factual or cultural knowledge across modalities, requiring inferential chaining rather than direct look-up or recall.

\textbf{Example}\\
\textit{General Question:} Associated with the community that celebrates Gudi Padwa and commonly seen hanging outside homes during festivals, which object made of cloth, neem leaves, and a copper vessel is shown in the image? \textit{(as referenced in the image)}\\
\textit{Answer:} Gudi
\end{tcolorbox}
\end{figure*}

\begin{figure*}[htbp]
\centering
\begin{tcolorbox}[colback=green!5!white,colframe=green!60!black,title=\textbf{Prompt 3: Analogy-Based Cultural Question}]
\textbf{Role:} Create a reasoning-based cultural question using analogy. The answer that is given below should be inferred by relating cultural equivalents or symbols. It is necessary to add a reference to the image of the cultural artifact like \textit{``as referenced in the image”}. DO NOT include any prefixes or labels like `Question:’. Return ONLY the rewritten question without any additional text.

\textbf{Justification:} This prompt targets abstract reasoning by requiring the model to draw symbolic or functional parallels between cultural entities—useful for assessing deeper conceptual understanding and metaphorical thinking.

\textbf{Example}\\
\textit{General Question:} Just as the red double-decker bus is iconic to London, which traditionally painted wooden vehicle, often seen in temple processions in Tamil Nadu, serves a similar symbolic role in South Indian culture? \textit{(as referenced in the image)}\\
\textit{Answer:} Temple chariot (Ratha)
\end{tcolorbox}
\end{figure*}

\end{document}